\pdfoutput=1

\documentclass[11pt]{article}

\usepackage[]{acl}

\usepackage{times}
\usepackage{latexsym}

\usepackage[T1]{fontenc}

\usepackage[utf8]{inputenc}

\usepackage{microtype}

\usepackage{algorithm}
\usepackage{algorithmic}
\usepackage{siunitx}    
\usepackage{relsize}
\usepackage{etoolbox}
\robustify\smaller

\usepackage{newfloat}
\usepackage{listings}
\floatstyle{ruled}
\newfloat{listing}{tb}{lst}{}
\floatname{listing}{Listing}

\usepackage{soul}
\usepackage{amsfonts}
\usepackage{multirow}

\usepackage{dsfont}
\usepackage{commath}

\usepackage{amsmath}
\usepackage{mathtools}

\usepackage{bbold}

\usepackage[utf8]{inputenc}
\usepackage{pgfplots}
\DeclareUnicodeCharacter{2212}{−}
\usepgfplotslibrary{groupplots,dateplot}
\usetikzlibrary{patterns,shapes.arrows}
\pgfplotsset{compat=newest}

\usepackage{placeins}
\usepackage{subcaption}
\usepackage{mwe}
\usepackage{pifont}
\usepackage{xcolor}
\usepackage{paralist}

\usepackage{booktabs}
\usepackage{relsize}
\usepackage{array}

\usepackage[capitalise]{cleveref}

\newcolumntype{V}{>{\smaller}l}

\newcommand{\class}[1]{\textsf{#1}\xspace}
\newcommand{\AAE}{\class{AAE}}
\newcommand{\SAE}{\class{SAE}}
\newcommand{\happy}{\class{happy}}
\newcommand{\sad}{\class{sad}}

\newcommand{\dataset}[1]{\textsc{#1}\xspace}
\newcommand{\Moji}{\dataset{Moji}}
\newcommand{\Bios}{\dataset{Bios}}

\newcommand{\method}[1]{\textsc{#1}\xspace}
\newcommand{\Standard}{\method{Standard}}
\newcommand{\INLP}{\method{INLP}}
\newcommand{\RW}{\method{RW}}
\newcommand{\DS}{\method{DS}}
\newcommand{\Adv}{\method{Adv}}
\newcommand{\DAdv}{\method{DAdv}}
\newcommand{\Gate}{\method{Gate}}
\newcommand{\BT}{\method{BTEO}}
\newcommand{\largeModel}{\method{Large}}

\newcommand{\adv}{\ensuremath{\text{adv}}}

\newcommand{\acc}{\ensuremath{\text{Accuracy}}}
\newcommand{\tpr}{\ensuremath{\text{TPR}}}

\newcommand{\GAP}{\ensuremath{\text{GAP}}\xspace}
\newcommand{\DTO}{\ensuremath{\text{DTO}}\xspace}

\usepackage{xspace}
\usepackage{adjustbox}

\usepackage{mathtools, caption}

\title{Balancing out Bias: Achieving Fairness Through Balanced Training 
}

\author{
Xudong Han$^{1}$ \qquad  Timothy Baldwin$^{1,2}$ \qquad    Trevor Cohn$^{1}$\\ 
$^{1}$The University of Melbourne \\ 
$^{2}$MBZUAI \\ 
\url{xudongh1@student.unimelb.edu.au}, \url{{tbaldwin,t.cohn}@unimelb.edu.au}
}

\begin{document}

\maketitle

\begin{abstract}
Group bias in natural language processing tasks manifests as disparities in system error rates across texts authorized by different demographic groups, typically disadvantaging minority groups. Dataset balancing has been shown to be effective at mitigating bias, however existing approaches do not directly account for correlations between author demographics and linguistic variables, limiting their effectiveness. To achieve Equal Opportunity fairness, such as equal job opportunity without regard to demographics, this paper introduces a simple, but highly effective, objective for countering bias using balanced training.
We extend the method in the form of a gated model, which incorporates protected attributes as input, and show that it is effective at reducing bias in predictions through demographic input perturbation, outperforming all other bias mitigation techniques when combined with balanced training.\footnote{Code available at \url{https://github.com/HanXudong/fairlib}}\end{abstract}

\section{Introduction}

Natural Language Processing (NLP) models have achieved extraordinary gains across a variety of tasks in recent years. 
However, naively-trained models often learn spurious correlations with other demographics and socio-economic factors~\citep{hendricks2018women, lu2018gender, bolukbasi2016man, park2018reducing}, leading to disparities across author demographics in contexts including coreference resolution, sentiment analysis, and hate speech detection~\citep{badjatiya2019stereotypical, zhao2018gender, li-etal-2018-towards, diaz2018addressing}.

Two popular approaches for mitigating such biases are: (1) balancing each demographic group in training, either explicitly via sampling~\citep{zhao2018gender, wang2019balanced} or implicitly via balancing losses for each group~\citep{hofler2005use, NEURIPS2020_07fc15c9}; and (2) removing demographic information from learned representations~\citep{li-etal-2018-towards, wang2019balanced,ravfogel-etal-2020-null, han2021diverse}.

While balancing methods have been shown to be successful, they have not been tested extensively in NLP. 
In this paper, we focus on Equal Opportunity fairness~(EO: \citet{hardt2016equality}), which requires non-discrimination across demographics within the “advantaged”
outcome labels, and adapt three balanced training approaches for debiasing.
In addition, we propose a new objective for balanced training, which can be used for proxy optimization of EO fairness.
We first provide a theoretical justification for our approach, and then conduct experiments on two benchmark datasets which show that our proposed objective is highly effective at achieving EO fairness while maintaining competitive accuracy.

Even when the training data is balanced, ignoring demographic-specific features can lead to bias~\citep{wang2019balanced,NEURIPS2020_07fc15c9}, due to differences in language use across demographics~\citep{hovy-2015-demographic}.
There is thus a fine line to be walked in terms of optimizing for linguistic variables associated with different demographic groups (potentially boosting overall model accuracy), and ensuring model fairness.

Inspired by work in domain adaptation on learning domain-specific representations that generalize across domains~\citep{NIPS2016_45fbc6d3, li-etal-2018-whats}, we propose a gated model, which incorporates author demographics as an \emph{input} to generate group-specific representations but also generalizes across demographic groups.
We show that when combined with instance reweighting during training, this technique leads to substantial bias reductions over leading debiasing techniques, typically with higher predictive accuracy.    
We also introduce a second means of bias reduction through tailoring gating coefficients of the trained model, which allows for fine-tuning of the accuracy--fairness trade-off.
Our experiments over two benchmark datasets for language debiasing show that our techniques are competitive with much more complex state-of-the-art methods for debiasing in situations where the demographic attribute is not known at test time, and provide substantial gains over the state-of-the-art when the protected attribute is observed.

\section{Balanced Training}
Despite their simplicity and versatility, balanced training approaches have received limited attention in prior work in NLP.
In this section, we propose a novel objective for balanced training, which we show to be a proxy for the EO.
We further review three balanced training approaches, discuss their objectives, and highlight their differences over our proposed method.

\subsection{Problem Formulation}
In this paper, we focus on bias mitigation for NLP classification tasks. Formally, we assume a dataset $\mathcal{D} = \{(x_{i}, y_{i}, g_{i})\}_{i=1}^{n}$ where $x_{i} \in X$ is a $d$-dimensional input text representation vector, $y_{i} \in Y$ denotes the main task label (e.g., sentiment), and $g_{i} \in G$ represents the private attribute associated with $x_{i}$ (e.g., author gender).

A standard model $M$ is trained to predict $Y$ given $X$, while debiasing methods generally aim to learn a model $M'$ that is fair \emph{wrt} $G$ by considering $X \times G$ together.

Let $\mathcal{X}$ be the task loss and $n$ be the number of observed instances in the dataset $\mathcal{D}$.
The overall empirical risk is written as $\mathcal{L} = \frac{1}{n}\sum_{i}\mathcal{X}(y_{i},\hat{y}_{i})$,
which can be rewritten as the aggregation of subsets:
$\mathcal{L}=\sum_{\text{y}} \sum_{\text{g} } \frac{n_{\text{y},\text{g}}}{n}\mathcal{L}_{\text{y},\text{g}},$
where $n_{\text{y},\text{g}}$ is the number of instances with target label $\text{y}$ and demographic attribute $\text{g}$, and $\mathcal{L}_{\text{y},\text{g}}$ is the empirical loss corresponding to the subset: $\mathcal{L}_{\text{y},\text{g}} = \frac{1}{n_{\text{y},\text{g}}}\sum_{i}\mathcal{X}(y_{i},\hat{y}_{i})\mathbb{1}(y_{i} = \text{y}, g_{i}=\text{g})$.

Furthermore, we use $*$ to denote marginalization, for example, $n_{*,\text{g}} = \sum_{\text{y}}n_{\text{y},\text{g}}$.
Let $p$ be the target label distribution, and $\tilde{p}$ be the empirical probability based on the training dataset.

\subsection{Fairness Measurement}
\emph{Equality of Opportunity} (EO) is widely used in previous work \citep{hardt2016equality, ravfogel-etal-2020-null, han-etal-2021-decoupling}, and measures the difference in true positive rate (TPR, aka Recall) across all groups, based on the notion that the positive outcome represents `advantage', such as getting a job or a loan.
Essentially, the difference (gap) in TPR reflects the degree to which different groups lack equal opportunity (with lower numbers indicating greater equity).

\subsection{Towards Equal Opportunity}
\label{sec:beq}

Without loss of generality, we illustrate with the binary case of $y \in \{T, F\}$ and $g \in \{0, 1\}$.
Recall that the equal opportunity metric is satisfied if a binary classification model has an equal positive prediction rate for the advantaged class. 
Assuming the advantaged class is $y=T$, the equal opportunity is measured by the TPR gap between protected groups, i.e., $\text{Recall}_{\text{g}=0}-\text{Recall}_{\text{g}=1}$.
Our proposed objective function for equal opportunity is:
\begin{equation*} 
\begin{split}
    \mathcal{L}^{EO} &= \frac{n_{T,*}}{n}\frac{1}{2}\sum_{\text{g}\in{\{0,1\}}}\mathcal{L}_{T,\text{g}}+\sum_{\text{g}\in{\{0,1\}}} \frac{n_{F,\text{g}}}{n}\mathcal{L}_{F,\text{g}}\\ 
    &=\sum_{\text{g}\in{\{0,1\}}} \frac{n_{T,\text{g}}}{n} \frac{n_{T,*}}{2n_{T,\text{g}}} \mathcal{L}_{T,\text{g}} +\sum_{\text{g}\in{\{0,1\}}} \frac{n_{F,\text{g}}}{n}\mathcal{L}_{F,\text{g}}\\
\end{split}
\end{equation*} 
Compared to the vanilla objective, the weights of instances with target label $T$  are adjusted. 
Specifically, the reweighting term $\frac{n_{T,*}}{2n_{T,\text{g}}}>1$ for the minority group, and $<1$ for the majority group.

\paragraph{From CE to \tpr}
Cross-entropy is an estimate of the \tpr\ at the mini-batch level when considering a subset of instances with the same target label.
Recall that the CE loss for binary classification of an instance is
$-[y_{i}\cdot \log(\hat{p}(y_{i})) + (1-y_{i})\cdot \log(1-\hat{p}(y_{i}))],$ where $\hat{p}(y_{i})$ is the predicted probability of $y_{i}$ being True.
Taking $y=T$ for a certain demographic group $\text{g}$ as an example, 
\begin{equation*}
    \begin{split}
        \mathcal{L}_{T,\text{g}} &=\frac{1}{n_{T,\text{g}}}\sum_{i}\mathcal{X}(y_{i},\hat{y}_{i})\mathbb{1}(y_{i} = T, g_{i}=\text{g}) \\
        &= -\frac{1}{n_{T,\text{g}}}\sum_{i} \log(\hat{p}(y_{i})) \mathbb{1}(y_{i} = T, g_{i}=\text{g}).
    \end{split}
\end{equation*}
Essentially, minimizing $\mathcal{L}_{T,\text{g}}$ is proportionate to maximizing the TPR of demographic group $\text{g}$. That is, at the minibatch level, $-\mathcal{L}_{T,\text{g}}$ is an estimator of $\log(p(\hat{y}=T|y=T, g=\text{g}))$, which is the log-\tpr\ of group $\text{g}$.
Given this, our proposed objective minimizes the \tpr\ gap by focusing on the log-\tpr\ difference across demographic groups.

\paragraph{Beyond binary labels~\&~demographic attributes}
Our proposed objective generalizes to higher ordering labels and demographic attributes trivially.
The equal opportunity metric was originally designed for binary classification, under the assumption of a single advantaged class $y=T$. 
To satisfy the multi-class target label case, we adjust the equal opportunity to consider the one-vs-all setting, and measuring the \tpr\ of each target class.
Our proposed objective then becomes $\sum_{\text{y}}\sum_{\text{g}} \frac{n_{\text{y},\text{g}}}{n} \frac{n_{\text{y},*}}{|G|\times n_{\text{y},\text{g}}} \mathcal{L}_{\text{y},\text{g}}$. 

\begin{figure}
    \centering
    \includegraphics[width=0.45\textwidth]{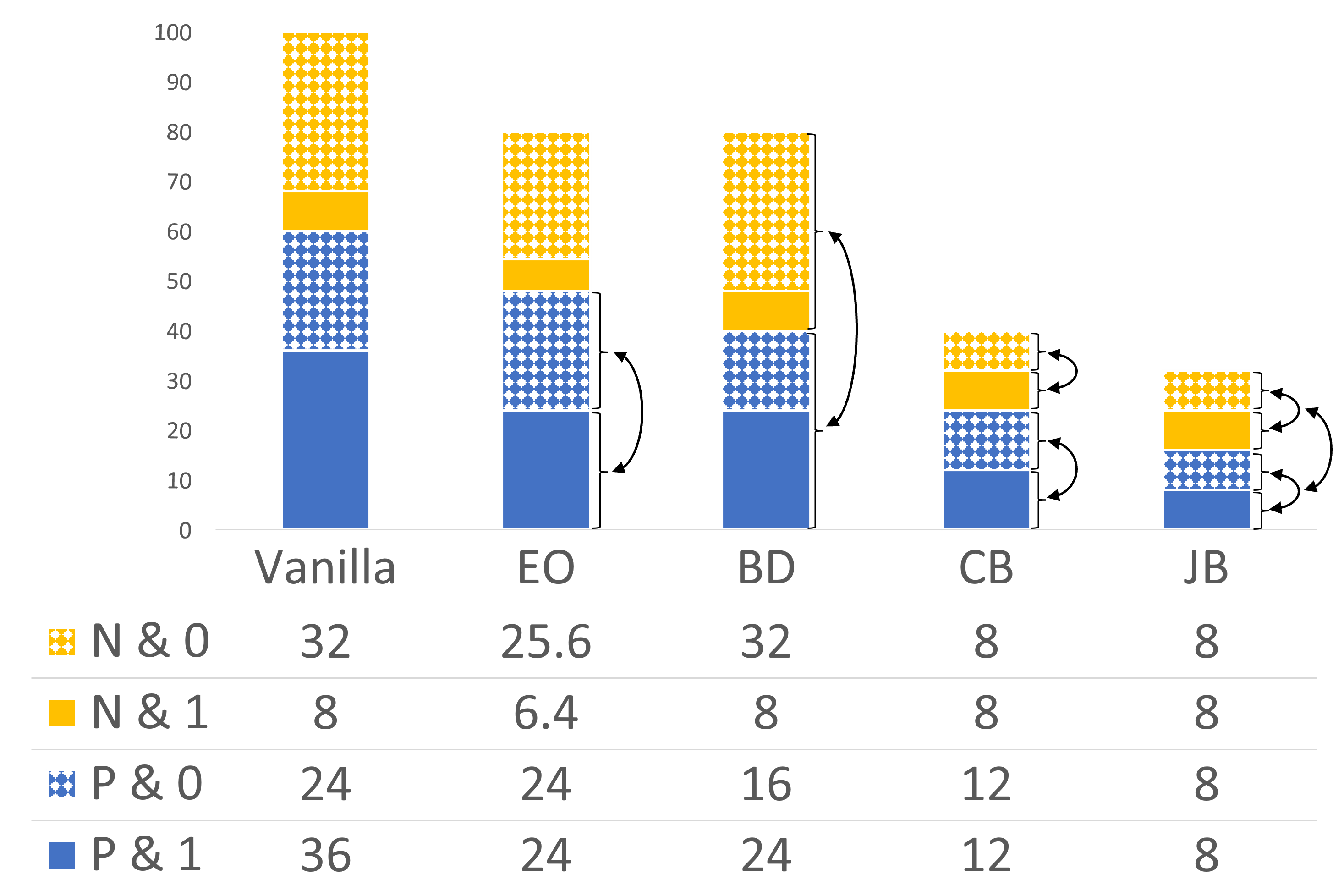}
    \caption{Balanced training through downsampling w.r.t.\ different objectives over a toy dataset for illustrative purposes. \textbf{N/P} = main task, \textbf{0/1} = protected attribute. The groups at either end of each arrow are resampled to be the same size. \textbf{EO} = our proposed objective in \cref{sec:beq}. \textbf{BD}, \textbf{CB}, and \textbf{JB} refer to balanced demographics, conditional balance, and joint balance, resp.\ (\cref{sec:previous_objectives}). }
    \label{fig:BT_intuition}
\end{figure}

\subsection{Balanced Training Objectives}
\label{sec:previous_objectives}
We now formally describe the objective functions of three established balanced training approaches, and discuss their applications.
We provide a toy example in Figure~\ref{fig:BT_intuition} to illustrate the differences between these objectives.
In~\cref{sec:mapping_of_objectives}, we provide more details about the mapping from previous work to these objectives in our framework.

\paragraph{Balanced Demographics (BD)} \citet{zhao2018gender} augment the dataset according to the demographic label distribution (making $p(G)$ uniform) for bias mitigation in the context of coreference resolution.
Although their gender-swapping approach is not directly applicable to our tasks, we adapt the general objective function as 
$\mathcal{L}^{G} =  \frac{1}{|G|} \sum_{\text{y}} \sum_{\text{g}} \frac{n_{\text{y}, \text{g}}}{n_{*,\text{g}}} \mathcal{L}_{\text{y},\text{g}}$,
where $|G|$ is the number of distinct labels of $G$.

Since $\mathcal{L}^{G}$ only encourages the model to equally focus on different demographic groups, it does not explicitly capture the correlation between $G$ and $Y$, and as a result, does not achieve Equal Opportunity fairness.

\paragraph{Conditional Balance (CB)} In a vision context, \citet{wang2019balanced} down-sample the majority demographic group within each class, so that on a per-class basis, it does not dominate the minority group (i.e.\ $p(G|Y)$ is uniform for all $Y$), giving the objective function:
$\mathcal{L}^{G|Y} =  \frac{1}{|G|} \sum_{\text{y}} \frac{n_{\text{y},*}}{n} \sum_{\text{g}} \mathcal{L}_{y,\text{g}}.$

This is the closest formulation to ours, as it also captures the conditional independence between $G$ and $Y$.
However, it captures both the TPR and TNR, while our method and EO fairness only focus on the TPR. 
In the multi-class target label case, our EO objective recovers the formulation of $\mathcal{L}^{G|Y}$.

\paragraph{Joint Balance (JB)} \citet{NEURIPS2020_07fc15c9} employ instance reweighting for structural data classification such that demographics and classes are jointly balanced,
leading to the objective:
$\mathcal{L}^{G,Y} =  \frac{1}{|G| \times |Y|} \sum_{\text{y}} \sum_{\text{g}} \mathcal{L}_{y,\text{g}}$.

JB can be treated as a combination of the classic long-tail learning objective and the CB objective ($p(G,Y)=p(G|Y)p(Y)$). 
On the one hand, JB is equivalent to CB when $Y$ has already been balanced, which is the case for the dataset \Moji (\cref{sec:dataset}), and CB is not a suitable objective for achieving EO fairness in this case.
On the other hand, when $Y$ is imbalanced, JB not only requires CB but also focuses more on long-tail main task classes, making it highly vulnerable to the size of minority groups.

\subsection{Achieving the Objective} 
In this paper, we focus on two ways of achieving the target objective: (1) instance reweighting, which manipulates the weight of each instance during training; and (2) down-sampling, which preprocesses the dataset before training.

Taking the joint balance (JB)
as an example, instance reweighting reweights each instance in inverse proportion to the frequency of the combination of its main label and demographic label, 
$\frac{1}{|\mathcal{D}|}\sum_{(x_{i}, y_{i}, g_{i}) \in \mathcal{D}} \tilde{p}^{-1}(G=g_{i},Y=y_{i}) \mathcal{X}(y_i, \hat{y}_i),$
where $\mathcal{X}$ is the task loss, and $\hat{y_i}$ denotes the model prediction given input text $x_{i}$. 

The other approach, down-sampling, sub-samples non-minority instances to derive a balanced training dataset, such that $\tilde{p}(g,y) = \frac{1}{|G| \times |Y|},\forall g \in G, y \in Y$. 
Specifically, let $\mathcal{D}_{y,g} = \{(x_{i}, y_{i}, g_{i})|y_{i} = y, g_{i}=g\}_{i=1}^{n}$ denote a subset for training.
We sample without replacement to get a target subset $\mathcal{D}_{y,g}^{*}$ such that $|\mathcal{D}_{y,g}^{*}| = \min \{ |\mathcal{D}_{y',g'}|, \forall y' \in Y, g' \in G \}$. 
The sampled subsets are merged to form the training set.

\section{Demographic Factors Improve Fairness}

Ignoring demographic-specific features can lead to bias even when the training data has been balanced \citep{wang2019balanced,NEURIPS2020_07fc15c9}.
Instead, as suggested by \citet{hovy-yang-2021-importance}, addressing demographic factors is essential for NLP to get closer to the goal of human-like language understanding, and increase fairness.
Our approach to dealing with this is, rather than removing demographic information, to use a gated model that uses demographic labels as input.

\begin{figure}[t]
    \centering
    \includegraphics[width=0.45\textwidth]{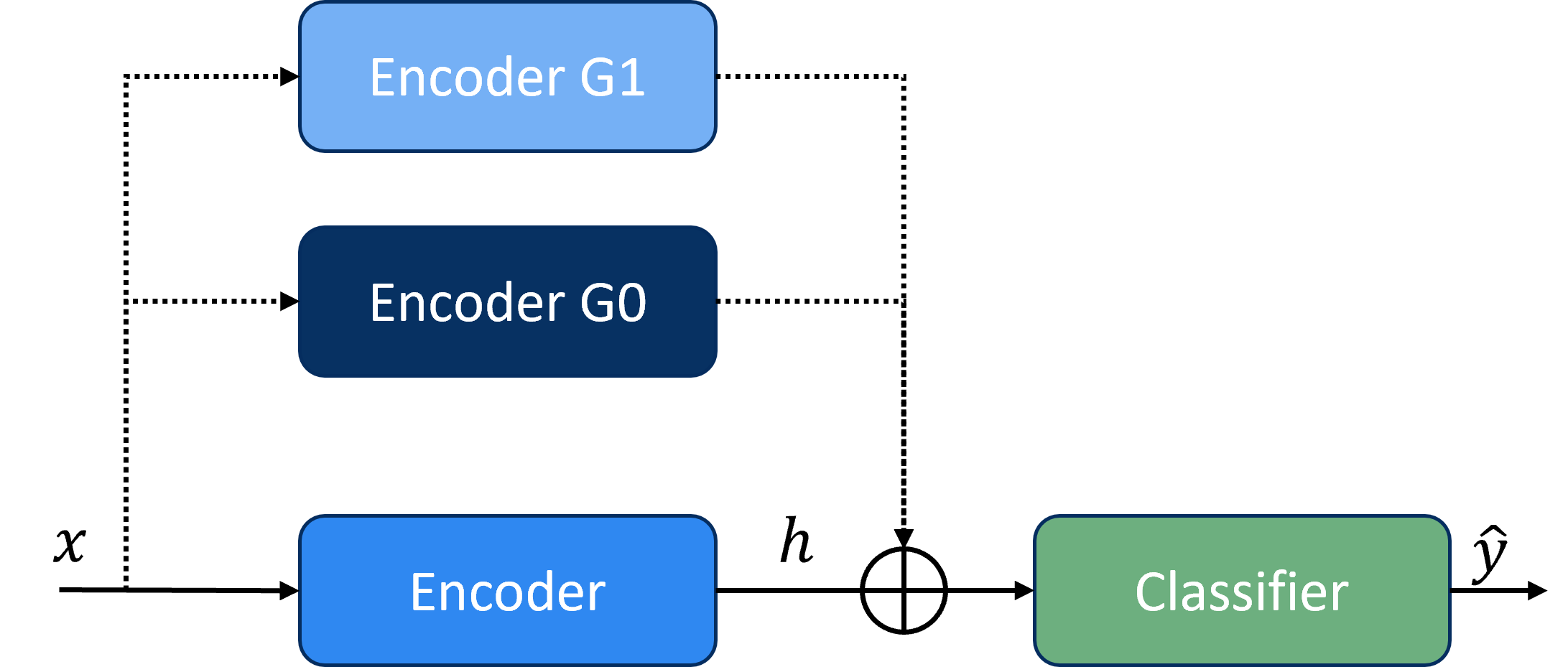}
    \caption{Gated model architecture. Given the input vector $x$, e.g.\ a text representation, the model has a shared encoder component and $|G|$ encoder components, one for each demographic group.}
    \label{fig:gated_model}
\end{figure}

As can be seen in Figure~\ref{fig:gated_model}, the gated model consists of $(1+|G|)$ encoders: one shared encoder, and a dedicated encoder for each demographic group in $G$.\footnote{Strictly speaking, it is possible to achieve a similar effect with $|G|$ encoders by merging one group with the shared encoder, and using post-hoc correction to separate out the general from the group-specific representation \citep{Kang:2020}.}  
Formally, let $E$ denote the shared encoder, $E_{j}$ denote the encoder for the $j$-th demographic group, $C$ denote the classifier, and $g_{i}$ be a 1-hot input such that $g_{i,j}$ is $1$ if the instance $(x_{i}, g_{i}, y_{i})$ belongs to the $j$-th group, and 0 otherwise. The prediction for an instance is:
$\hat{y}_{i} = C(h_{i}^{s}, h_{i}^{g})$,
where $h_{i}^{s} = E(x_{i})$ and $h_{i}^{g} = \sum_{j=1}^{|G|}g_{i,j}E_{j}(x_{i})$.
The two inputs are concatenated and input to the classifier $C$.

Intuitively, the shared encoder learns a general representation, while each group-specific encoder captures group-specific representations.

Our setting differs from other debiasing methods in that we assume the demographic attribute is available at training and prediction time, while techniques such as adversarial training~\citep{li-etal-2018-towards} and INLP~\citep{ravfogel-etal-2020-null} only require the attribute for training. 
This richer input allows for more accurate predictions, courtesy of the demographic-specific encoder, but limits applicability at test time.
For better applicability, we also relax this requirement by replacing demographic factors with a non-informative prior in \cref{sec:soft_averaging}.

\section{Experimental Results}
In this section we first introduce our experimental settings, and then report and discuss our results. 
In~\cref{sec:reproducibility}, we provide detailed settings for reproducing our experiments.

\subsection{Evaluation Metrics}
Following \citet{ravfogel-etal-2020-null}, we use overall accuracy as the performance metric, and the separation criterion to measure fairness in the form of TPR \GAP and TNR \GAP: the true positive rate and true negative rate differences between demographic groups. For both \GAP metrics, smaller is better, and a perfectly fair model will achieve 0.
For multi-class classification tasks, we follow \citet{ravfogel-etal-2020-null} in reporting the quadratic mean (RMS) of TPR \GAP over all classes. 
In a binary classification setup, TPR and TNR are equivalent to the TPR of the positive and negative classes, respectively, so we employ the RMS TPR \GAP in this case also.

Throughout this paper, we report accuracy and \GAP results as mean values $\pm$ standard deviation over the test set, averaged across five independent runs with different random seeds.

In contrast to single-objective evaluation, evaluation of fairness approaches generally reports both fairness and performance at the same time. Typically, there is no single method that achieves both the best performance and fairness, making comparison between different fairness methods difficult. 
This problem has been widely studied in the literature on \emph{multi-objective learning}~\citep{marler2004survey}.
For ease of comparison between approaches, we adopt the \emph{compromise solution}~\citep{salukvadze1971concerning} to fairness evaluation, and introduce `distance to the optimum' (DTO).
Specifically, the \emph{compromise solution} aims to minimize the difference between the candidate point and a \emph{utopia point}. 
In our case, the candidate points are ordered pairs $(\acc,\GAP)$, denoting the accuracy and fairness of debiasing methods, and the \emph{utopia point} (optimum) represents the hypothetical system which achieves the highest-achievable accuracy and fairness for the dataset., i.e., $(\text{max}(\acc),\text{min}(\GAP))$.
Following~\citet{vincent1981optimality,vincent1983game}, \DTO is calculated as the Euclidean distance between the optimum and the results for a given method. Lower is better for this statistic, with a minimum of 0.
In \cref{sec:calculation_of_dto}, we provide a more detailed explanation of \DTO, including a step-by-step example calculation.

In addition, we are also interested in the efficiency of the different debiasing approaches and report each method's average training time.\footnote{Testing on Titan X and RTX 3090, all models have roughly identical inference time.}

\subsection{Dataset}
\label{sec:dataset}

Following \citet{ravfogel-etal-2020-null}, we conduct experiments over two NLP classification tasks --- sentiment analysis (\Moji) and biography classification (\Bios) --- using the same dataset splits.
In~\cref{sec:data_distribution}, we provide analysis of the dataset distribution.

{\bf \Moji}: This sentiment analysis dataset was collected by \citet{blodgett-etal-2016-demographic}, and contains tweets that are either African American English (AAE)-like or Standard American English (SAE)-like.
Each tweet is annotated with a binary `race' label (based on language use: either \AAE or \SAE), and a binary sentiment score determined by (redacted) emoji contained in it.

{\bf \Bios}: The second task is biography classification~\citep{de2019bias}, where biographies were scraped from the web, and annotated for binary gender and 28 classes of profession.

\subsection{Models}
We first implement a ``\Standard'' model on each dataset, without explicit debiasing. On the \Moji dataset, we follow~\citet{ravfogel-etal-2020-null} in using DeepMoji~\citep{felbo2017} as the encoder to get 2304d representations of input texts. 
\citet{ravfogel-etal-2020-null} and \citet{subramanian2021} used uncased BERT-base~\citep{devlin2019bert} as their \Standard model for the \Bios dataset, taking the `CLS' token as the source of a fixed text representation, without further fine-tuning.
However, we found that taking the average of all contextualized token embeddings led to an accuracy improvement of 1.4\% and \GAP fairness improvement of 2.4\%.
Given this, we use 768d `AVG' representations extracted from the pretrained uncased BERT-base model.

\subsection{Balanced Training Approaches}

\begin{table}[t!]
    \centering
    \begin{adjustbox}{max width=\linewidth}
    \sisetup{
    round-mode = places,
    }%
    \begin{tabular}{
    l
    S[table-format=2.1, round-precision = 1]@{\,\( \pm \)\,}S[table-format=1.2, round-precision = 1,table-number-alignment = left]
    S[table-format=2.1, round-precision = 1]@{\,\( \pm \)\,}S[table-format=1.2, round-precision = 1,table-number-alignment = left]
    r
    }

    \toprule
    \bf Model & \multicolumn{2}{c}{\bf Accuracy$\uparrow$}    &   \multicolumn{2}{c}{\bf \GAP$\downarrow$} & \bf \DTO $\downarrow$ \\ \midrule
    \Standard                       &   82.3   & 0.02  &   15.96   &   0.51    &   0.093\\[0.9ex]
    \textbf{BD}~\citep{zhao2018gender}   &   82.3   & 0.01  &   15.63   &   0.19    &   0.089 \\
    \textbf{JB}~\citep{NEURIPS2020_07fc15c9}     &   74.71   & 0.28  &   7.35    &   0.30    &   0.092 \\
    \textbf{EO}   &   79.42   & 0.05  &   9.66    &   0.64    &   {\bf 0.043}\\
    \bottomrule
    \end{tabular}
    \end{adjustbox}
    \caption{Results for balanced training methods on the
      \Bios test set. \textbf{EO}: our proposed objective in \cref{sec:beq}. \textbf{BD} and \textbf{JB} are baselines from \cref{sec:previous_objectives}. \textbf{Bold} = best trade-off.
    }
    \label{tab:bios_full_Balanced_results}
\end{table}

Since the \Moji dataset has been artificially balanced for main task and demographic labels, balanced training based on $p(g)$ makes no difference, and moreover, the results for $p(g|y)$ and $p(g,y)$ will be identical. 
Given this, we focus on the \Bios dataset for comparing different balanced training objectives.\footnote{As \Bios is a multi-class classification task and our proposed approach generalizes to \textbf{BD} in this case, there is no need to include \citet{wang2019balanced} in our comparison.}

\cref{tab:bios_full_Balanced_results} shows the results of balanced training using the different objectives.
Compared to the \Standard model, balanced training with different objectives are all able to reduce bias, and the objective proposed by \citet{NEURIPS2020_07fc15c9} achieves the lowest TPR GAP.
However, in terms of accuracy--fairness trade-off, our proposed approach outperforms all other models, which is not surprising as it is designed to achieve better equal opportunity fairness.
Based on these results, hereafter, we only report balanced training with our proposed EO objective (\BT).

\begin{table*}[ht!]
\renewrobustcmd{\bfseries}{\fontseries{b}\selectfont}
\centering
\begin{adjustbox}{max width=\linewidth}
\sisetup{
round-mode = places,
}%
\begin{tabular}{
ll
S[table-format=3.1, round-precision = 1]@{\,\( \pm \)\,}S[table-format=1.1, round-precision = 1,table-number-alignment = left]
S[table-format=3.1, round-precision = 1]@{\,\( \pm \)\,}S[table-format=1.1, round-precision = 1,table-number-alignment = left]
S[table-format=1.3, round-precision = 3]
S[table-format=3.1, round-precision = 1]
S[table-format=3.1, round-precision = 1]@{\,\( \pm \)\,}S[table-format=1.1, round-precision = 1,table-number-alignment = left]
S[table-format=3.1, round-precision = 1]@{\,\( \pm \)\,}S[table-format=1.1, round-precision = 1,table-number-alignment = left]
S[table-format=1.3, round-precision = 3]
S[table-format=3.1, round-precision = 1]
}

\toprule
& & \multicolumn{6}{c}{\bf\Moji} & \multicolumn{6}{c}{\bf\Bios} \\
\cmidrule(lr){3-8}\cmidrule(lr){9-14}

\bf Method & \bf Model         & \multicolumn{2}{c}{\bf Accuracy$\uparrow$}     &\multicolumn{2}{c}{\bf \GAP$\downarrow$} & \multicolumn{1}{c}{\bf \DTO$\downarrow$ } & \multicolumn{1}{c}{\bf Time$\downarrow$ } & \multicolumn{2}{c}{\bf Accuracy$\uparrow$}     &\multicolumn{2}{c}{\bf \GAP$\downarrow$} & \multicolumn{1}{c}{\bf \DTO$\downarrow$ } & \multicolumn{1}{c}{\bf Time$\downarrow$ } \\ 
\midrule

\multirow{4}{*}{Baselines} & \Standard      & 71.59    &     0.05    & 30.96  &  0.28 &   0.261011311 & 1.00 &   82.27 & 0.02   & 15.96  & 0.51  & 0.110176176 & 1.00 \\
 & \INLP        & 68.54     & 1.05      & 33.83  & 3.91 &    0.29983885 & 14.00 &    70.54  & 0.54  & 6.69 & 0.86  & 0.144886815 & 6.33 \\
 & \Adv           & 74.25    &     0.39  & 22.19  &  3.72 & 0.162737646 & 36.06 & 81.09 & 0.08 & 12.70 & 0.28  & 0.07684874 & 1.26 \\
 & \DAdv  & 74.52&     0.28 & 18.48 &  1.96 & 0.123 & 109.40 & 81.07 & 0.05 & 12.56 & 0.28 &  0.076 & 2.35 \\
\cmidrule(lr){2-8} \cmidrule(lr){9-14}
\multirow{4}{*}{Ours} & \BT  &  74.01&     0.21    & 21.48 &  0.38 & 0.155  & 0.79 & 79.42 & 0.05    & 9.66 & 0.64  & 0.057 & 0.67 \\
& \BT+\Gate$^{*}$ & 74.89&   0.19  & 13.77 &  0.30 & 0.072 & 0.82 & 79.44 & 0.07 & 9.20 & 0.24  & \bfseries{0.053} & 0.67 \\ [1ex]
& \Gate$^{*}$  & 64.82& 0.13 & 65.20 &  0.88 & 0.639775981 & 1.0 & 82.37 & 0.06 &   19.23 & 0.32    &   0.144067797 & 1.0 \\
& \Gate$_\text{RMS}^{\text{soft}}$ $^{*}$  & 73.54    &   0.224130  &  7.06    &  0.270208 & \bfseries{0.019} & 1.0 & 80.54 & 0.072115    &   11.08 & 0.319592  & 0.063 & 1.0\\


\bottomrule
\end{tabular} 
\end{adjustbox}
\caption{
Results over the sentiment analysis (\Moji) and biography classification (\Bios) tasks. \DTO\ is measured by the normalized Euclidean distance between each model and the ideal model, and lower is better. \textbf{Bold} = best trade-off within category. 
Normalized time is reported relative to \Standard, which takes 35 secs and 16 mins for \Moji and \Bios, respectively. The reported times are the average times divided by that of \Standard. $^{*}$ indicates that the model requires the demographic attribute at test time.}
\label{table:main_results}
\end{table*}

\subsection{Main Results}

We report results over the sentiment analysis and biography classification tasks in Table~\ref{table:main_results}.
The baseline models are: {\bf\Standard}, which is a naively-trained MLP classifier; {\bf\INLP}~\citep{ravfogel-etal-2020-null}, which removes demographic information from text representations through iterative nullspace projection; {\bf\Adv}~\citep{li-etal-2018-towards,wang2019balanced, zhao2019conditional}, which performs protected information removal through adversarial training; and {\bf\DAdv}~\citep{han2021diverse}, which also uses adversarial training but with multiple adversaries subject to an orthogonality constraint, and represents the current best baseline models.

On the \Moji dataset, compared to the \Standard model, \BT simultaneously increases accuracy and mitigates bias, leading to results competitive with \Adv and better than \INLP.
Although \BT does not outperform the best baseline models \DAdv, it leads to performance--fairness trade-offs that are competitive with the other debiasing methods.

On the \Bios dataset, \BT again leads to performance--fairness trade-offs that outperform the baseline methods. 
However, different to the \Moji dataset, \BT does not further improve accuracy, improving fairness by 5.3\% absolute at the cost of 2.9\% accuracy.

In terms of training time, existing debiasing methods (esp.\ \DAdv on \Moji) incur a substantial overhead, while balanced training is much more frugal: around 1.3 times faster (because of the reduction in training data volume).

In addition to evaluating \BT, we also combine \Gate with \BT, which achieves a better performance--fairness balance, as shown in Table~\ref{table:main_results}.
This is consistent with our argument that, rather than removing demographic information, properly used demographic factors can further reduce biases.
Indeed, the \BT+\Gate consistently outperforms the current best baseline models model \DAdv on both datasets.

In~\cref{sec:combinations}, we also show that \BT can be combined with \DAdv and \INLP, leading to better bias mitigation.

\subsection{Gated Model}
If the training dataset is imbalanced and contains spurious correlations between task labels and demographic attributes, a naively trained model will learn and possibly amplify dataset biases.
The \Gate model, with its explicit conditioning and group-specific encoding, will be particularly vulnerable to bias.

Table~\ref{table:main_results} shows that, on both datasets, the \Gate model increases the accuracy but amplifies bias (e.g., GAP of 65 on \Moji): as it uses demographic information directly to make predictions, it is highly vulnerable to bias in the training dataset. 

Intuitively, the only objective of \Gate training is standard cross-entropy loss, which has been shown to lead to bias amplification under imbalanced training without regularization. 
The gate components explicitly rely on demographic information, and thus become a strong indicator of predictions due to spurious correlations between the main task label and demographic labels in the training set.

Balanced training approaches act as regularizers in preventing the model from learning and amplifying spurious correlations in training.

\subsection{Soft Averaging}
\label{sec:soft_averaging}
Although the gated model naturally requires the demographic attribute at test time, we also evaluate a condition where this is not available. 
Instead, we take a Bayesian approach by evaluating $p(y|x) = \sum_g p(g) p(y|g,x)$, where we can control the prior explicitly.
For example, under a uniform demographic attribute prior, we simply average the predictions $p(y|x,g)$ and $p(y|x,\neg g)$.
This Bayesian approach can be approximated by soft averaging, whereby the activation of all demographic-specific encoders are uniformly averaged inside the model,
 i.e., $g_{i,j} = \frac{1}{|G|}$, rather than selecting only one in the standard gated model (i.e., $g_{i,:}$ is 1-hot).

\begin{figure}[t!]
    \centering
     \begin{subfigure}[b]{0.225\textwidth}
         \centering
         \includegraphics[width=\textwidth]{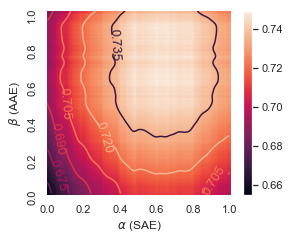}
         \caption{\Moji Accuracy}
         \label{fig:moji_valid_acc}
    \end{subfigure}
    \hfill
    \begin{subfigure}[b]{0.225\textwidth}
        \centering
        \includegraphics[width=\textwidth]{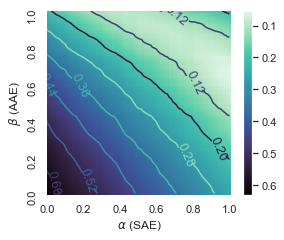}
        \caption{\Moji \GAP}
        \label{fig:moji_valid_rms}
    \end{subfigure}
    \hfill
    \\[2ex]
     \begin{subfigure}[b]{0.225\textwidth}
         \centering
         \includegraphics[width=\textwidth]{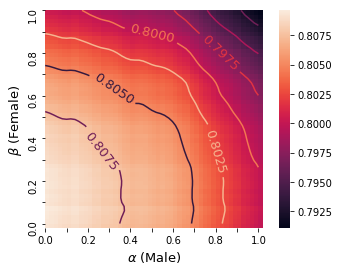}
         \caption{\Bios Accuracy}
         \label{fig:bios_valid_acc}
    \end{subfigure}
    \hfill
    \begin{subfigure}[b]{0.225\textwidth}
        \centering
        \includegraphics[width=\textwidth]{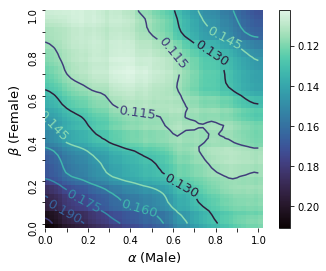}
        \caption{\Bios \GAP}
        \label{fig:bios_valid_rms}
    \end{subfigure}
    \hfill
    \\
    \caption{Accuracy and \GAP of $\alpha$ and $\beta$ settings for \Moji and \Bios. The axes refer to the propensity to change the gold group in gating the encoder components, and the bottom left point $\alpha=\beta=0$ is the \Gate model using true demographic inputs. Lighter shading denotes better performance. 
    }
    \label{fig:gate_soft_perturbation}
\end{figure}

When the protected attribute is \textbf{observed} at test time the soft averaging method may still prove useful, which we use as a means for fine-tuning the balance between accuracy and bias.
Figure~\ref{fig:gate_soft_perturbation} shows an example for the prior fine-tuning.
Specifically, we consider non-uniform encoder averaging conditioned on the gold protected attribute, $g^\ast$.
Let $\alpha$ and $\beta$ denote to what extent the 1-hot labels are manipulated according to the value of $g^\ast$ as 0 and 1 respectively, leading to
the soft labels $\begin{bmatrix} \alpha & 1-\alpha \end{bmatrix}$ and $\begin{bmatrix} 1-\beta & \beta \end{bmatrix}$.
I.e., the two specific encoders are weighted by either $\alpha$ and $1-\alpha$, or $1-\beta$ and $\beta$, respectively, according to the value of $g^\ast$.
Values of $\alpha, \beta < 0.5$ mean the protected label is (softly) preserved, while values $>0.5$ mean the label is flipped.

In cases where the model is biased towards or against a demographic group, it may be advantageous to use these two additional parameters to correct for this bias, by disproportionately using the other group's encoder.

We now employ the Bayesian ``soft averaging'' approach to gating, and mitigating bias at inference time. Note that this does not involve retraining the model, as the soft averaging happens at test time.

Figure~\ref{fig:gate_soft_perturbation} shows accuracy and \GAP results from tuning the coefficients on development data for the basic \Gate model.
The results show that $\alpha=\beta=0.5$ is a reasonable default setting, however gains may be possible for non-uniform prior settings.

To demonstrate the power of adjusting these parameters, we take the trained \Gate model, and then optimize $\alpha$ and $\beta$ over the development set, and report the corresponding results on the test set.
We select the parameter values that achieve the lowest development \GAP, provided accuracy is above a threshold.
\footnote{
The [$\alpha, \beta$] values are [0.64, 0.99] and [0.38, 0.72] over \Moji and \Bios, respectively, We also experimented with adjusting the gating coefficients for \Gate + \BT, in which case there was no benefit from using non-zero $\alpha$ or $\beta$.
}
The results are reported in Table~\ref{table:main_results}, under \Gate$_\text{RMS}^{\text{soft}}$.
On the \Moji dataset, our results show that \Gate with soft averaging consistently outperforms the \Standard and \Gate models without balanced training.
In terms of \GAP, the model is substantially better than all other models, while remaining competitive in terms of accuracy.
The \Bios dataset is noisier, meaning there are bigger discrepancies between the development and test datasets.
However, we achieve a good performance--fairness trade-off at a level comparable to the much more complex \INLP and \DAdv models.

\section{Analysis}

\subsection{Reweighting vs.\ Down-sampling}
\begin{table}[t!]
    \centering
    \begin{adjustbox}{max width=\linewidth}
    \sisetup{
    round-mode = places,
    }%
    \begin{tabular}{
    lr
    S[table-format=2.1, round-precision = 1]@{\,\( \pm \)\,}S[table-format=1.2, round-precision = 1,table-number-alignment = left]
    S[table-format=2.1, round-precision = 1]@{\,\( \pm \)\,}S[table-format=1.2, round-precision = 1,table-number-alignment = left]
    r
    }

    \toprule
    \bf Model   &   \bf Size & \multicolumn{2}{c}{\bf Accuracy$\uparrow$}    &   \multicolumn{2}{c}{\bf \GAP$\downarrow$} & \bf \DTO $\downarrow$ \\ \midrule
    \Standard    & 257k    &   82.27 & 0.02   & 15.96  & 0.51 & 0.093\\[0.9ex]
    \RW + \textbf{BD}    & 257k     &    82.28  & 0.01  & 15.63 & 0.19 & 0.089 \\
    \RW + \textbf{JB}    & 257k  &   74.71 & 0.28    &   7.35 & 0.30 & 0.092\\
    \RW + \textbf{EO}    & 257k    &    75.66  & 0.17  & 13.93 & 0.37 & 0.107\\[0.9ex]
    \DS + \textbf{BD}    & 237k  &    82.07 & 0.07   &  15.87 & 0.32 & 0.092\\
    \DS + \textbf{JB}    & 5k & 66.08 & 0.09    & 10.85 & 0.36 &0.200\\
    \DS + \textbf{EO}    & 37k & 79.42 & 0.05    & 9.66 & 0.64 &\textbf{0.043}\\
    \bottomrule
    \end{tabular}
    \end{adjustbox}
    \caption{Results for balanced training methods on the
      \Bios test set. ``RW'' = instance reweighting; ``DS'' = dataset
      down-sampling; and ``Size'' = the number of instances in the training dataset. \textbf{Bold} = best trade-off.
    }
    \label{tab:bios_full_4RW_results}
\end{table}

Table~\ref{tab:bios_full_4RW_results} shows the results of the naively-trained MLP model (``\Standard'') and six balanced-training methods, all based on the same architecture as \Standard. Corpus down-sampling (``\DS'') removes instances from majority groups and thus leads to less training data and overall lower accuracy than instance reweighting (``\RW'').

When using \textbf{BD} as the objective, both \RW and \DS perform similarly to the \Standard model, as the overall gender distribution is quite balanced, which can also be seen in the size of the training data for \DS + \textbf{BD}.
Both \RW + \textbf{JB} and \RW + \textbf{EO} reduce bias and performance, but \RW + \textbf{JB} outperforms \RW + \textbf{EO} in terms of the performance--fairness trade-off, in that \RW + \textbf{JB} achieves similar performance but substantially better fairness (6.6\% absolute improvement in \GAP).
However, \textbf{JB} is not as effective as \textbf{EO} when combined with \DS, due to the big drop in the volume of training data.

\subsection{\Gate vs. \ \largeModel}
\label{sec:num_parameters}
Compared to \Standard, \Gate involves  more model parameters, and an important question is whether the gains of \Gate models are simply because of the larger parameter space.

To explore this question, we conduct experiments over \Moji by controlling the number of parameters of \Standard and compare it with \Gate models. 
Specifically, we employ a larger \Standard model, namely \largeModel, which has more hidden units within each hidden layer, leading to roughly the same number of parameters as the \Gate model.

Table~\ref{tab:results_with_parameters} shows the results for \Standard \largeModel and \BT with \largeModel. 
Comparing \Standard and \BT with the \largeModel versions, it can be seen that increasing the number of parameters does not increase performance or fairness.

Despite \BT+\Gate having fewer parameters than \BT \largeModel, it achieves a substantial improvement in terms of fairness, confirming our argument that \Gate is superior to existing models.


\begin{table}[t!]
    \centering
    \begin{adjustbox}{max width=\linewidth}
    \begin{tabular}{lrcc}
    \toprule
         \textbf{Model}  &   \textbf{Parameters}  & \textbf{\acc} & \textbf{\GAP} \\\hline
          \Standard & 782402 & 71.6 & 31.0 \\
          \BT & 782402 & 74.0   &   21.5   \\
          \BT+\Gate & 2346002 & 74.9    &   13.8    \\
          \midrule
          \Standard \largeModel & 2887202 &  71.7    &   31.4    \\
          \BT \largeModel & 2887202 & 73.9    &   20.9    \\
    \bottomrule
    \end{tabular}
    \end{adjustbox}
    \caption{Results over \Moji. \largeModel modes have larger hidden sizes to achieve similar number of parameters of our proposed \Gate model. }
    \label{tab:results_with_parameters}
\end{table}

\section{Related Work}

\paragraph{Fairness}
Much work on algorithmic fairness has focused on group fairness, i.e., disparities in error rates across groups defined by protected attributes, such as gender, age, or race. 
Many criteria have been proposed for group fairness, such as statistical parity \citep{dwork2012fairness} and equal opportunity~\citep{hardt2016equality}.
Broadly speaking, fairness can be classified into three categories: independence, separation, and sufficiency \citep{barocas-hardt-narayanan}, with the most recent work addressing separation criteria, i.e, potential correlations between main task labels and protected attributes. 

\paragraph{Mitigating bias}
Many approaches for bias mitigation haven been proposed recently, including removing protected information form hidden representations~\citep{li-etal-2018-towards,ravfogel-etal-2020-null,han2021diverse}, preprocessing data to remove bias~\citep{zhao2018gender, vanmassenhove-etal-2018-getting, saunders-byrne-2020-reducing}, modifying the training algorithm~\citep{badjatiya2019stereotypical}, and post-hoc correction~\citep{hardt2016equality}.

In the context of NLP, the best results have been achieved through protected information removal.
Iterative nullspace projection (\citet{ravfogel-etal-2020-null}) takes hidden representations and projects them onto the nullspace of the weights of a linear classifier for each protected attribute.
The classifier training and projection are carried out over multiple iterations to more comprehensively remove protected information.

Another popular approach is adversarial training, which jointly optimizes the removal of sensitive information and main task performance, through the incorporation of adversarial discriminator(s) to identify protected attributes from the hidden representations~\citep{li-etal-2018-towards,elazar-goldberg-2018-adversarial, wang2019balanced}.
Differentiated adversarial learning~(\citet{han2021diverse}) uses an ensemble of adversaries for each protected attribute, subject to an orthogonality constraint.

\section{Conclusions and Future Work}
This paper proposed the adoption of balanced training approaches to mitigate bias, and demonstrated their effectiveness relative to existing methods, as well as their ability to further enhance existing methods.
We also proposed a gated model based on demographic attributes as an input, and showed that while the simple version was highly biased, with a simple Bayesian extension at inference time, the method was highly effective at mitigating bias.

For future work, it is important to consider settings where there are multiple protected attributes, such as author age, gender, and ethnicity. 
A simple extension would be to treat $G$ as being \emph{intersectional classes}, defined as the Cartesian product of the multiple demographic groups. 
E.g., $k$ binary groups would result in $2^k$ intersectional classes.

\section*{Ethical Considerations}
This work aims to advance research on bias mitigation in NLP. Although our proposed method requires access to training datasets with protected attributes, this is the same data assumption that is made by other related work such as adversarial training and INLP, and our target is to make fair predictions at inference time. One limitation of methods in this area, including ours, is the difficulty of training and evaluating when we don't have access to demographic attributes. To avoid user harm, we only use attributes which users have self identified in our experiments. Moreover, our proposed method is able to make fairer predictions either with or without demographic information. All data in this study is publicly available and used under strict ethical guidelines.

\section*{Limitations}
(1) We only investigate bias mitigation over English tasks. However, our method is naturally generalized to other languages.

(2) Many of debiasing methods assume that protected attributes are available in the dataset. While we acknowledge that protected attributes are normally unavailable, we follow previous work's setup and only use self-identified attributes in experiments. for future work is incorporating models that can debias with less protected labels in order to deal with real-world data, such as decoupled adversarial training~\citep{han-etal-2021-decoupling} and adversarial instance reweighting~\citep{NEURIPS2020_07fc15c9}.

(3) For both \INLP and \DAdv, we follow experimental setup from the original papers.
However, the \texttt{fairlib}~\citep{han2022fairlib} --- which is presented after this work was done --- recently show that both methods can obtain better results with a larger budget for hyperparameter fine-tuning.

\bibliography{custom}
\bibliographystyle{acl_natbib}

\clearpage

\appendix

\section{Dataset distribution}
\label{sec:data_distribution}

\subsection{\Moji}
This training dataset has been artificially balanced according to demographic and task labels, but artificially skewed in terms of race--sentiment combinations, as follows: \AAE--\happy = 40\%, \SAE--\happy = 10\%, \AAE--\sad = 10\%, and \SAE--\sad = 40\%.
We used the train, dev, and test splits from \citet{han2021diverse} of 100k/8k/8k instances, respectively. 

\subsection{\Bios}
Since the data is not directly available, in order to construct the dataset, we re-scrape the data with the scripts of \citet{ravfogel-etal-2020-null}, leading to a dataset with 396k biographies, which we randomly split into train (65\%), dev (10\%), and test (25\%).

\begin{figure}[ht]
    \centering
    \includegraphics[width=0.45\textwidth]{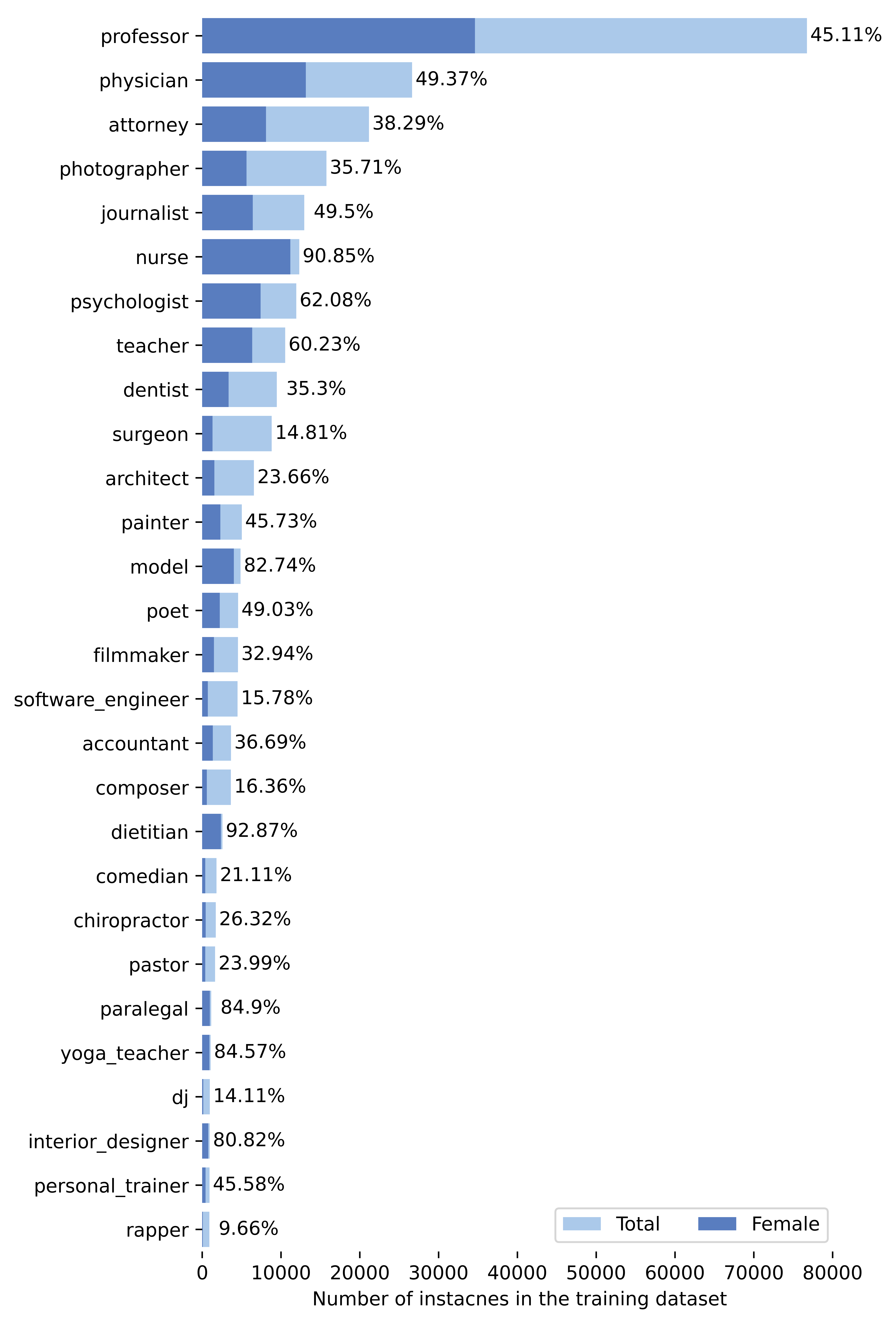}
    \caption{Bios dataset statistics.}
    \label{fig:bios_statistic}
\end{figure}

Figure~\ref{fig:bios_statistic} shows the statistic of the \Bios dataset. Each row corresponds to a profession, including the total number of instances and number of female instances. 
Besides, each profession is also annotated with the percentage of female instances.
There are slight discrepancies in the dataset composition due to data attrition: the original dataset~\citep{de2019bias} had 399k instances, while 393k were collected by~\citet{ravfogel-etal-2020-null}.

\section{Reproducibility}
\label{sec:reproducibility}

\subsection{Models}
For \INLP~\cite{ravfogel-etal-2020-null}, we take the fixed \Standard model for the given dataset, and iteratively train a linear classifier and perform nullspace projection over the learned representation.
For the other baseline models --- \Adv and \DAdv --- we jointly train the adversarial discriminators and classifier.
In order to ensure a fair comparison, we follow~\citet{han-etal-2021-decoupling} in using a model consisting of the same fixed-parameter encoder as ours followed by a trainable 3-layer MLP.

\subsection{Hyperparameter Tuning}
All approaches proposed in this paper share the same hyperparameters as the standard model.
Hyperparameters are tuned using grid-search, in order to maximise accuracy for the standard model, and to minimise the fairness \GAP for debiasing methods, subject to the accuracy exceeding a given threshold.
The accuracy threshold is chosen to ensure the selected model achieves comparable performance to baseline methods, defined as up to 2\% less than best baseline accuracy. 
Taking \RW as an example, the best baseline accuracy on the \Bios development dataset is 75.7\% and accordingly the (development) accuracy threshold is set to 73.7\%; among models in the hyperparameter search space that exceed this threshold, we take the model with minimum \GAP.
We report test results for the selected models.

In terms of the baseline models, both \DAdv and \INLP have additional hyperparameters:
for \DAdv these are the weight of the adversarial loss, which controls the performance--fairness trade-off; the number of sub-adversaries; and the weight of the difference loss, to better remove demographic information;
while \INLP also has a trade-off hyperparameter, the number of null-space projection iterations, and other hyperparameters related to linear attackers and classifiers.

The trade-off hyperparameter makes such models more flexible in performing model selection.
However, it also requires manual selection for better trade-offs, and different strategies have been introduced.
For example, \INLP manually selects the model at a iteration where the accuracy is minimally damaged while the fairness improves greatly.
Similar manual selection for better trade-offs is also required for \Adv and \DAdv, but the strategies proposed in the original papers are slightly different to one another, and are also task-specific.

In order to reproduce previous methods, we follow the original paper in setting the accuracy threshold, and then tuning hyperparameters for the best fairness.

For the \Adv and \DAdv models, following the work of~\citet{han2021diverse},  we tune extra hyperparameters separately, such as the trade-off hyperparameter, while using the same shared hyperparameters to the selected base models.
Similarly, the number of iterations for the \INLP model is tuned once other hyperparameters have been fixed.

\begin{table*}[ht!]
    \begin{adjustbox}{max width=\linewidth}

    \centering
    \begin{tabular}{ccccccccc}
        \toprule 
        & & \multicolumn{7}{c}{\bf Best assignment} \\
        \cmidrule(lr){1-2} \cmidrule(lr){3-9}
        \textbf{Hyperparameter} & \textbf{Search space} & \bf \Standard & \bf \Adv & \bf \DAdv & \bf \DS & \RW & \DAdv + \DS & \DAdv + \RW  \\
        \cmidrule(lr){1-2} \cmidrule(lr){3-9}
        number of epochs & - & \multicolumn{7}{c}{100}\\
        \cmidrule(lr){1-2} \cmidrule(lr){3-9}
        patience & - & \multicolumn{7}{c}{10}\\
        \cmidrule(lr){1-2} \cmidrule(lr){3-9}
        encoder & - & \multicolumn{7}{c}{DeepMoji~\citep{felbo2017}}\\
        \cmidrule(lr){1-2} \cmidrule(lr){3-9}
        embedding size & - & \multicolumn{7}{c}{2304}\\
        \cmidrule(lr){1-2} \cmidrule(lr){3-9}
        hidden size & - & \multicolumn{7}{c}{300}\\
        \cmidrule(lr){1-2} \cmidrule(lr){3-9}
        number of hidden layers & \emph{choice-integer}[1, 3] & \multicolumn{7}{c}{2}\\
        \cmidrule(lr){1-2} \cmidrule(lr){3-9}
        batch size & \emph{loguniform-integer}[64, 2048] & 1024 & 1024 & 1024 & 512 & 1024 & 512 & 1024\\
        \cmidrule(lr){1-2} \cmidrule(lr){3-9}
        output dropout & \emph{uniform-float}[0, 0.5] & 0.4 & 0.4 & 0.4 & 0.5 & 0.5 & 0.2 & 0.1\\
        \cmidrule(lr){1-2} \cmidrule(lr){3-9}
        optimizer & - & \multicolumn{7}{c}{Adam~\citep{kingma:adam}}\\
        \cmidrule(lr){1-2} \cmidrule(lr){3-9}
        learning rate  & \emph{loguniform-float}[$10^{-6}$, $10^{-1}$] & $3\times10^{-5}$ & $3\times10^{-5}$ & $3\times10^{-5}$ & $10^{-5}$ & $10^{-4}$ & $3\times10^{-5}$ & $3\times10^{-4}$\\
        \cmidrule(lr){1-2} \cmidrule(lr){3-9}
        \textbf{l}earning \textbf{r}ate \textbf{s}cheduler & - & \multicolumn{7}{c}{reduce on plateau}\\
        \cmidrule(lr){1-2} \cmidrule(lr){3-9}
        \textbf{LRS} patience & - & \multicolumn{7}{c}{2 epochs}\\
        \cmidrule(lr){1-2} \cmidrule(lr){3-9}
        \textbf{LRS} reduction factor & - & \multicolumn{7}{c}{0.5}\\
        \cmidrule(lr){1-2} \cmidrule(lr){3-9}
        \Adv loss weight & \emph{loguniform-float}[$10^{-4}$, $10^{2}$] & - & $10^{-0.1}$ & $10^{-0.1}$ & - & - & $10^{0.2}$ & $10^{0.0}$\\
        \cmidrule(lr){1-2} \cmidrule(lr){3-9}
        \Adv hidden size & \emph{loguniform-integer}[64, 1024] & - & 256 & 256 & - & - & 256 & 256\\
        \cmidrule(lr){1-2} \cmidrule(lr){3-9}
        number of adversaries & \emph{choice-integer}[1, 8] & - & 1 & 3 & - & - & 3 & 3\\
        \cmidrule(lr){1-2} \cmidrule(lr){3-9}
        \DAdv loss weight & \emph{loguniform-float}[$10^{-5}$, $10^{5}$] & - & - & $10^{3.7}$ & - & - & $10^{2}$ & $10^{2.6}$\\
        \bottomrule
    \end{tabular}
    \end{adjustbox}
    \caption{Search space and best assignments on the \Moji dataset}
    \label{tab:moji_hyperparameter}
\end{table*}

\begin{table*}[ht!]
    \begin{adjustbox}{max width=\linewidth}

    \centering
    \begin{tabular}{ccccccccc}
        \toprule 
        & & \multicolumn{7}{c}{\bf Best assignment} \\
        \cmidrule(lr){1-2} \cmidrule(lr){3-9}
        \textbf{Hyperparameter} & \textbf{Search space} & \bf \Standard & \bf \Adv & \bf \DAdv & \bf \DS & \RW & \DAdv + \DS & \DAdv + \RW  \\
        \cmidrule(lr){1-2} \cmidrule(lr){3-9}
        number of epochs & - & \multicolumn{7}{c}{100}\\
        \cmidrule(lr){1-2} \cmidrule(lr){3-9}
        patience & - & \multicolumn{7}{c}{10}\\
        \cmidrule(lr){1-2} \cmidrule(lr){3-9}
        encoder & - & \multicolumn{7}{c}{uncased BERT-base~\citep{devlin2019bert}}\\
        \cmidrule(lr){1-2} \cmidrule(lr){3-9}
        embedding size & - & \multicolumn{7}{c}{768}\\
        \cmidrule(lr){1-2} \cmidrule(lr){3-9}
        embedding type & \emph{choice}\{\textbf{`CLS'}, \textbf{`AVG'}\} & \multicolumn{7}{c}{\textbf{`AVG'}}\\
        \cmidrule(lr){1-2} \cmidrule(lr){3-9}
        hidden size & - & \multicolumn{7}{c}{300}\\
        \cmidrule(lr){1-2} \cmidrule(lr){3-9}
        number of hidden layers & \emph{choice-integer}[1, 3] & \multicolumn{7}{c}{2}\\
        \cmidrule(lr){1-2} \cmidrule(lr){3-9}
        batch size & \emph{loguniform-integer}[64, 2048] & 512 & 128 & 128 & 128 & 256 & 256 & 512\\
        \cmidrule(lr){1-2} \cmidrule(lr){3-9}
        output dropout & \emph{uniform-float}[0, 0.5] & 0.5 & 0.3 & 0.2 & 0.3 & 0.5 & 0.2 & 0.4\\
        \cmidrule(lr){1-2} \cmidrule(lr){3-9}
        optimizer & - & \multicolumn{7}{c}{Adam~\citep{kingma:adam}}\\
        \cmidrule(lr){1-2} \cmidrule(lr){3-9}
        learning rate  & \emph{loguniform-float}[$10^{-6}$, $10^{-1}$] & $3\times10^{-3}$ & $10^{-3}$ & $10^{-3}$ & $10^{-3}$ & $3\times10^{-5}$ & $3\times10^{-3}$ & $3\times10^{-4}$\\
        \cmidrule(lr){1-2} \cmidrule(lr){3-9}
        \textbf{l}earning \textbf{r}ate \textbf{s}cheduler & - & \multicolumn{7}{c}{reduce on plateau}\\
        \cmidrule(lr){1-2} \cmidrule(lr){3-9}
        \textbf{LRS} patience & - & \multicolumn{7}{c}{2 epochs}\\
        \cmidrule(lr){1-2} \cmidrule(lr){3-9}
        \textbf{LRS} reduction factor & - & \multicolumn{7}{c}{0.5}\\
        \cmidrule(lr){1-2} \cmidrule(lr){3-9}
        \Adv loss weight & \emph{loguniform-float}[$10^{-8}$, $10^{2}$] & - & $10^{-2.3}$ & $10^{-2.3}$ & - & - & $10^{-2.8}$ & $10^{-5}$\\
        \cmidrule(lr){1-2} \cmidrule(lr){3-9}
        \Adv hidden size & \emph{loguniform-integer}[64, 1024] & - & 256 & 256 & - & - & 256 & 256\\
        \cmidrule(lr){1-2} \cmidrule(lr){3-9}
        number of adversaries & \emph{choice-integer}[1, 8] & - & 1 & 3 & - & - & 3 & 3\\
        \cmidrule(lr){1-2} \cmidrule(lr){3-9}
        \DAdv loss weight & \emph{loguniform-float}[$10^{-5}$, $10^{5}$] & - & - & $10^{2}$ & - & - & $10^{3}$ & $10^{3 .3}$\\
        \bottomrule
    \end{tabular}
    \end{adjustbox}
    \caption{Search space and best assignments on the \Bios dataset}
    \label{tab:bios_hyperparameter}
\end{table*}

\begin{table*}[ht!]
\renewrobustcmd{\bfseries}{\fontseries{b}\selectfont}
\centering
\begin{adjustbox}{max width=\linewidth}
\sisetup{
round-mode = places,
}%
\begin{tabular}{
ll
S[table-format=3.1, round-precision = 1]@{\,\( \pm \)\,}S[table-format=1.1, round-precision = 1,table-number-alignment = left]
S[table-format=3.1, round-precision = 1]@{\,\( \pm \)\,}S[table-format=1.1, round-precision = 1,table-number-alignment = left]
S[table-format=1.3, round-precision = 3]
S[table-format=3.1, round-precision = 1]
S[table-format=3.1, round-precision = 1]@{\,\( \pm \)\,}S[table-format=1.1, round-precision = 1,table-number-alignment = left]
S[table-format=3.1, round-precision = 1]@{\,\( \pm \)\,}S[table-format=1.1, round-precision = 1,table-number-alignment = left]
S[table-format=1.3, round-precision = 3]
S[table-format=3.1, round-precision = 1]
}
\toprule
& & \multicolumn{6}{c}{\bf\Moji} & \multicolumn{6}{c}{\bf\Bios} \\
\cmidrule(lr){3-8}\cmidrule(lr){9-14}

\bf Method & \bf Model         & \multicolumn{2}{c}{\bf Accuracy$\uparrow$}     &\multicolumn{2}{c}{\bf \GAP$\downarrow$} & \bf \DTO$\downarrow$ & \bf Time$\downarrow$ & \multicolumn{2}{c}{\bf Accuracy$\uparrow$}     &\multicolumn{2}{c}{\bf \GAP$\downarrow$} & \bf \DTO$\downarrow$ & \bf Time$\downarrow$ \\ 
\midrule

\multirow{4}{*}{Baselines} & \Standard      & 71.59    &     0.05    & 30.96  &  0.28 &   0.261011311 & 1.00 &   82.27 & 0.02   & 15.96  & 0.51  & 0.110176176 & 1.00 \\
 & \INLP        & 68.54     & 1.05      & 33.83  & 3.91 &    0.29983885 & 14.00 &    70.54  & 0.54  & 6.69 & 0.86  & 0.144886815 & 6.33 \\
 & \Adv           & 74.25    &     0.39  & 22.19  &  3.72 & 0.162737646 & 36.06 & 81.09 & 0.08 & 12.70 & 0.28  & 0.07684874 & 1.26 \\
 & \DAdv  & 74.52&     0.28              & 18.48 &  1.96 & \bfseries{0.123} & 109.40 & 81.07 & 0.05 & 12.56 & 0.28 &  \bfseries{0.076} & 2.35 \\
\cmidrule(lr){2-8} \cmidrule(lr){9-14}
\multirow{2}{*}{Balance} & \DS  &  71.90&    0.05   & 23.24  &  0.20 & 0.177872903  & 0.54 & 79.42 & 0.05    & 9.66 & 0.64  & \bfseries{0.057} & 0.33 \\
& \RW  &  74.01&     0.21    & 21.48 &  0.38 &    \bfseries{0.155} & 1.03 & 74.71 & 0.28   &  7.35 & 0.30  & 0.095372157 & 1.01 \\
\cmidrule(lr){2-8} \cmidrule(lr){9-14}
\multirow{3}{*}{Gate} & \Gate  & 64.82&    0.13 & 65.20 &  0.88 &   0.639775981 & 1.04 & 82.37 & 0.06    &   19.23 & 0.32  & 0.144067797 & 1.01 \\
& \Gate + \DS   & 72.49&   0.03  & 16.33 &  0.72 & 0.104086108 & 0.56 & 79.44 & 0.07 & 9.20 & 0.24  & \bfseries{0.053} & 0.33 \\
& \Gate + \RW  & 74.89&   0.19  & 13.77 &  0.30 & \bfseries{0.072} & 1.08 & 74.89 & 0.22    & 7.12 & 0.18  & 0.092396001 & 1.02 \\
\cmidrule(lr){2-8} \cmidrule(lr){9-14}
\multirow{3}{*}{Bayesian} 
& \Gate$_\text{0.5}^{\text{soft}}$ & 72.68    &     0.215697    &  30.231639 &  0.282511 & 0.250383281 & 1.04 &   80.820438 & 0.072115    &   11.612711 & 0.319592  & 0.066459204 & 1.01 \\[0.5ex]
& \Gate$_\text{Acc}^{\text{soft}}$ & 74.83&     0.215082    &  20.29 &  0.290920  & 0.142094539 &  1.04 &   81.13 & 0.093250    &   19.83 & 0.359092  & 0.15124881  & 1.01 \\[0.5ex]
& \Gate$_\text{RMS}^{\text{soft}}$  & 73.54    &   0.224130  &  7.06    &  0.270208 & \bfseries{0.019} & 1.04 &   80.54 & 0.072115    &   11.08 & 0.319592  & \bfseries{0.063} & 1.01 \\[0.5ex]
\cmidrule(lr){2-8} \cmidrule(lr){9-14}
\multirow{4}{*}{Combination} & \DAdv + DS & 72.21 & 0.16 & 14.33 & 0.20 & \bfseries{0.085} & 72.14 & 79.25 & 0.07    & 9.89 & 0.22 &  \bfseries{0.059} & 2.27 \\
& \INLP + \DS  & 72.14 & 1.60 & 18.43 & 3.05 & 0.127251166 & 6.31 & 73.18 & 0.61    & 5.91  & 0.82  & 0.111695704 & 1.31 \\
& \DAdv + \RW  &  74.64&     0.13   & 18.91 &  0.30 & 0.127081435 & 108.21 & 74.09 & 0.20     & 7.24 & 0.39  & 0.102144188 & 3.04 \\
& INLP + \RW & 72.28 & 1.88 & 15.68 & 3.07 & 0.098867012 & 13.91 & 73.58 & 0.62    & 5.57 & 0.74 & 0.106796117 & 6.34 \\
\bottomrule
\end{tabular} 
\end{adjustbox}
\caption{
Results over the sentiment analysis (\Moji) and biography classification (\Bios) tasks. Trade-offs are measured by the normalized Euclidean distance between each model and the ideal model, and lower is better. \textbf{Bold} = best trade-off within category. 
Training time is reported relative to \Standard, which takes 35 secs and 16 mins  for \Moji and \Bios, respectively.}
\label{table:full_results}
\end{table*}

\subsection{Training Details}
We conduct all our experiments on a Windows server with a 16-core CPU (AMD Ryzen Threadripper PRO 3955WX), two NVIDIA GeForce RTX 3090s with NVLink, and 256GB RAM.

\subsubsection{\Moji}
For all baseline models, we follow the method of~\citet{han2021diverse}.
Specifically, we train the \Standard model for 100 epochs with the Adam optimizer~\citep{kingma:adam}, learning rate of $3\times10^{-5}$, and batch size of 1024.
For \Adv, the main model is jointly trained together with adversaries which are implemented as 3-layer MLP, and the weight of adversarial loss is 0.8.
For each iteration (epoch) of the main model, an adversary is trained for 60 epochs, keeping the checkpoint model that performs best on the dev set.
Three sub-adversaries are employed by the \DAdv, with the difference losss weight of $10^{3.7}$.
For \INLP, logistic regression models are used for both identifying null-space to the demographic information at each iteration, and making the final predictions given debiased hidden representations.
Since the number of iterations in \INLP is highly affected by the random seed at each run, we re-select it at each iteration.

As for our models, the \DS model is trained with the learning rate of $10^{-5}$ and batch size of 512; the \RW is trained with the learning rate of $10^{-4}$ and batch size of 1024; and the \Gate is trained with the the set of hyperparameters to the base model.

\subsubsection{\Bios}
Models are trained with similar hyperparameters as models on the \Moji dataset. 
We thus only report main differences for each of them: the \Standard model is trained with the batch size of 512 and learning rate of $3\times10^{3}$; \DS models are trained with the batch size of 128 and learning rate of $10^{-3}$, and \RW models are trained with the batch of 256 and learning rate of $3\times10^{-5}$.

We train the \Adv model with the adversarial loss weight of $10^{-2.3}$, learning rare for adversarial training of $10^{-1}$, learning rate of $10^{-3}$, and batch size of 128.
The \DAdv is trained with same setting as the \Adv, excepting the difference loss weight of $10^{2}$.
For details of the assignment of other hyperparameters and hyperparameter searching space, refer to Supplementary Materials.

\section{The calculation of \DTO}
\label{sec:calculation_of_dto}

For ease of comparison between approaches, we introduce `distance to the optimum' (DTO), a single metric to incorporate accuracy and \GAP into a single figure of merit, which is calculated by: (1) converting \GAP to $1 - \GAP$ (denoted as fairness; higher is better); (2) normalizing each of accuracy and fairness, by dividing by the best result for the given dataset (i.e., highest accuracy and fairness); and (3) calculating the Euclidean distance to the point $(1,1)$, which represents the hypothetical system which achieves highest accuracy and fairness for the dataset.
Lower is better for this statistic, with minimum 0.

We calculate \DTO based on all results shown in Table~\ref{table:full_results}.
Taking the DAdv model on the Moji dataset for example, the trade-off is calculated as follows:
\begin{enumerate}
    \item Find the best accuracy and fairness (1-GAP) separately; i.e., 74.9 (\Gate + \RW) and 92.9 (\Gate$_\text{RMS}^{\text{soft}}$), resp.
    \item  Normalize the accuracy and fairness metric of \DAdv, resulting in $0.995=\frac{74.5}{74.9}$ and $0.877=\frac{81.5}{92.9}$.
    \item Calculate the Euclidean distance between $(1,1)$ and $(0.995, 0.877)$, giving  
    $0.123$.
\end{enumerate}

\section{Training time estimation}
Given that the training time is affected by factors such as batch size, hidden size, and learning rate, to perform a fair comparison between different models, we estimate the training time of a model based on hyperparameter tuning results, over a shared search space of base hyperparameters (i.e., the hyperparameters related to the standard model), with any other approach-specific hyperparameters fixed. 

\section{Mapping of previous objectives}
\label{sec:mapping_of_objectives}

As for the jointly balance, \citet{NEURIPS2020_07fc15c9} state that ``In addition to vanilla inverse probability weighting (IPW), we also report results for an IPW variant with inverse probabilities computed jointly over protected-features and class-label reported as IPW(S+Y)''. It clearly shows that instance from group $g$ with class $y$ is weighted by $\frac{1}{n_{y,g}}$. Adding this weight to the unweighted loss function leads to $\sum_{y}\sum_{g}\frac{n_{y,g}}{n}\frac{1}{n_{y,g}}\mathcal{L}_{y,g}$, which exactly the objective that is shown in our paper.

In terms of the conditionally balance, in the section `Alternative Data Splits' of \citet{wang2019balanced}, they obtain the dataset by resampling such that the number of occurrences of men with label y and of women with label y is close, i.e., the size of different demographic groups are almost identical for each class-label. This is equivalent to assign equal weights to difference demographic groups within each class-label, i.e., $\sum_{y}\sum_{g}\frac{n_{y,g}}{n}\frac{n_{y,*}}{n_{y,g}}\mathcal{L}_{y,g}$.

Finally, for the balanced demographics, \citet{zhao2018gender} balance the distribution of demographic groups which reweights instance of a demographic group inversely to its proportion, leading a weight $\frac{n}{n_{*,g}}$. As a result, the final objective is $\sum_{y}\sum_{g}\frac{n_{y,g}}{n}\frac{n}{n_{*,g}}\mathcal{L}_{y,g}$.

\section{Further extensions}

\subsection{Combining balanced training with benchmark methods}
\label{sec:combinations}

The baseline methods \INLP and \DAdv as presented above were used in a manner consistent with their original formulation, i.e., without balanced training.
An important question is whether balanced training might also benefit these methods. 
It is trivial to combine downsampling with \INLP and \DAdv, as the method simply prunes the training dataset, but does not impact the training objective.
To combine instance reweighting with \DAdv, we modify the training objective such that the cross-entropy term is scaled by $\tilde{p}^{-1}$, while leaving the adversarial term unmodified, i.e., solve for $\min_{M}\max_{A}\sum_{(x_{i}, y_{i}, g_{i}) \in \mathcal{D}} \tilde{p}^{-1}\mathcal{X}(y_i, \hat{y}_i) - \lambda_{\adv} \mathcal{X}(g, \hat{g})$.
For \INLP, we simply train a \BT model, and then iteratively perform INLP linear model training and nullspace projection over the learned representations.

Results are presented in the final section of Table~\ref{table:full_results} (``Combination''), and indicate that the combined methods appreciably outperform both the standalone demographic removal methods and balanced training approaches, without extra training time cost. That is, demographic information removal and balanced training appear to be complementary.

\subsection{Balancing toward anti-stereotyping}
\label{sec:anti-stereotyping}

As shown in Table~\ref{table:main_results}, even with \DS or \RW balancing, the model still shows biases in its predictions.
We conduct preliminary experiments on \Moji with \RW and \DS, while controlling for stereotyping skew in training using values for 0.8 to 0.2.
In standard rebalancing we use as target 0.5, which describes a balanced situation. 
A larger skew $>0.5$ will amplifying stereotyping, and $<0.5$ describes a different type of stereotyping operating in the opposite direction.
Balancing towards a 0.4 training skew leads to the best test results, with an accuracy of 71.7\% and \GAP of 11.8\% for \DS, and accuracy of 74.5\% and \GAP of 11.3\% for \RW.
Comparing to the corresponding values in Table~\ref{table:main_results} (rows Balance \DS and \RW, for \Moji), both results show a substantial reduction in \GAP.

This idea is related to existing reweighting approaches in long-tail learning. For example, \citet{Cui:2019} infer the effective number of samples which group each instance with its neighbours within a small region instead of using all data points, and reweight the loss of each class inversely proportional to the effective number of samples.
We leave this further exploration of this line of research to future work.

We also experiment with \Gate+\RW and \Gate+\DS with a 0.4 training skew, however, the gated model does not show the same behaviour, as it just amplifies the training biases.
This implies that, for the gated model, balanced training can help remove spurious correlations between protected attributes and main task labels, which is similar in nature to the effects of adversarial training.

\section{Comparison with MoE}

One line of work that is similar to our gated model is mixture of experts~\citep{ma2018modeling,fedus2021switch}.
Technically, the gated model is similar to the MoE model in the sense that an expert can be largely aligned with a group-specific encoder in our model. However, there are several key differences: (1) instead of making independent predictions by each expert, our group-invariant encoder acknowledges the shared patterns across demographic groups; (2) MoE employs an extra softmax gating network to mix experts' predictions, while our method does an argmax based on group labels; and (3) we use the group-specific information jointly together with the group-invariant encoder' outputs for making the final predictions while the MoE model has one output layer for each expert.

\end{document}